\documentclass[11pt]{article} 
\usepackage{rldmsubmit,palatino}
\usepackage{graphicx}

\usepackage{amssymb}
\usepackage{hyperref}

\newcommand{\orcid}[1]{\raisebox{1pt}{\href{https://orcid.org/#1}{\includegraphics[height=10pt]{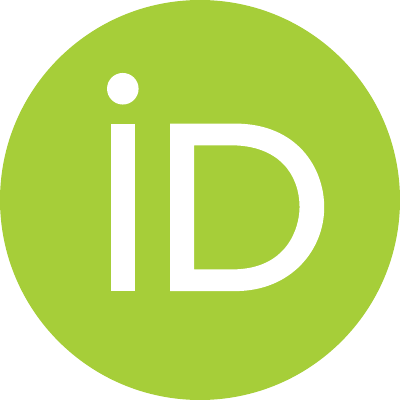}}}}
\usepackage[%
style=apa,%
backend=biber,%
citestyle=authoryear,%
natbib=true,%
uniquename=false,%
useprefix=true%
]{biblatex}
\title{Learning Relative Return Policies With Upside-Down Reinforcement Learning}

\author{
Dylan R.~Ashley \thanks{Correspondence to \href{mailto:dylan.ashley@idsia.ch}{\texttt{dylan.ashley@idsia.ch}}}\enspace$^{1,2,3}$ \orcid{0000-0001-6148-8802} \\
\And
Kai Arulkumaran $^{4,5}$ \\
\AND
J{\"{u}}rgen Schmidhuber $^{1,2,3,6,7}$ \\
\And
Rupesh Kumar Srivastava $^{7}$ \\
\AND
{\normalfont $^1$ The Swiss AI Lab IDSIA, Lugano, Switzerland} \\
$^2$ Universit{\`{a}} della Svizzera italiana (USI), Lugano, Switzerland \\
$^3$ Scuola universitaria professionale della Svizzera italiana (SUPSI), Lugano, Switzerland \\
$^4$ ARAYA Inc., Tokyo, Japan \\
$^5$ Imperial College London, London, UK \\
$^6$ AI Initiative, King Abdullah University of Science and Technology (KAUST), Thuwal, Saudi Arabia \\
$^7$ NNAISENSE, Lugano, Switzerland \\
}

\begin{document}

\maketitle

\begin{abstract}

Lately, there has been a resurgence of interest in using supervised learning to solve reinforcement learning problems. Recent work in this area has largely focused on learning command-conditioned policies. We investigate the potential of one such method---upside-down reinforcement learning---to work with commands that specify a desired relationship between some scalar value and the observed return. We show that upside-down reinforcement learning can learn to carry out such commands online in a tabular bandit setting and in CartPole with non-linear function approximation. By doing so, we demonstrate the power of this family of methods and open the way for their practical use under more complicated command structures.

\end{abstract}


\acknowledgements{
This work was supported by the European Research Council (ERC, Advanced Grant Number 742870) and the Swiss National Supercomputing Centre (CSCS, Project s1090). We also thank both the NVIDIA Corporation for donating a DGX-1 as part of the Pioneers of AI Research Award and IBM for donating a Minsky machine.
}

\startmain 

\section{Introduction}
\label{sec:introduction}

Artificial neural networks in their current incarnation are better suited to solving supervised learning problems than they are for solving reinforcement learning problems. Recently, a family of techniques based on upside-down reinforcement learning~\citep{schmidhuber2019reinforcement,srivastava2019training} has been proposed, which solve RL problems by framing them as supervised learning problems. These techniques all focus on directly learning a command-conditioned policy; they explicitly learn a mapping from states and commands to actions. Already, these methods have had remarkable success in solving offline RL problems~\citep{kumar2019reward,chen2021decision,janner2021reinforcement}, but still struggle to achieve competitive results in online reinforcement learning problems.

This work investigates the learnability of \textit{morethan} commands: commands that specify a goal in the form of a scalar value, a horizon, and the desired relation between the given scalar and the observed return under the given horizon. Here we show that these commands are learnable online with traditional UDRL in simple settings. By doing so, we demonstrate the practical potential of the flexibility of commands offered by the UDRL framework. We hope that this flexibility can be leveraged in the future to empower these methods to render them competitive in complicated online and continual learning problems.

\section{Related Work}
\label{sec:related_work}

The idea of leveraging iterated supervised learning to solve reinforcement learning dates back to at least the work on reward-weighted regression by \citet{peters2007reinforcement}, who brought the earlier work of \citet{dayan1997using} to the domain of operational space control and RL. However, \citet{peters2007reinforcement} only looked at the immediate-reward RL setting. This was extended to the episodic setting separately by \citet{wierstra2008episodic} and then by \citet{kober2011policy}. \citet{wierstra2008episodic} went even further and also extended RWR to partially observable Markov decision processes, whereas \citet{kober2011policy} applied it to motor learning in robotics. Separately, \citet{wierstra2008fitness} extended RWR to perform fitness maximization for evolutionary methods. \citet{hachiya2009efficient,hachiya2011reward} later found a way of reusing old samples to improve RWR's sample complexity. Much later, \citet{peng2019advantage} modified RWR to produce an algorithm for off-policy RL, using deep neural networks as function approximators.

Upside-down reinforcement learning as a command-conditioned method of using SL for RL emerged in \citet{schmidhuber2019reinforcement} and \citet{srivastava2019training}, with \citet{ghosh2021learning} afterwards introducing a similar idea in a multi-goal context. \citet{kumar2019reward} applied UDRL to offline RL and sometime later \citet{chen2021decision} and then \citet{janner2021reinforcement} showed the potential of the UDRL framework to be competitive in this context when paired with the Transformer architecture of \citet{vaswani2017attention}. \citet{furuta2021generalized} generalized this Transformer variant to solve a broader class of problems.

\section{Background}
\label{sec:background}

Reinforcement learning considers an agent receiving rewards through interacting with an environment. RL is usually modeled as a \textit{Markov decision process} where, at each step $t$, the agent observes the current state of the environment $s_t \in \mathit{S}$, selects an action $a_t \in \mathit{A}$, and consequently receives a reward $r_{t + 1} \in \mathbb{R}$. We say that the rule an agent follows to select actions is its \textit{policy}, which we write as $\pi : \mathit{S} \times \mathit{A} \rightarrow \mathbb{R}$ where $\pi(s, a)$ is the probability of the agent taking action $a$ when the environment is in state $s$. The objective is typically to find a policy that maximizes the sum of the temporally-discounted rewards---known as the return.

Upside-down reinforcement learning breaks the RL problem in two, using commands as an intermediary. Specifically, it divides the agent into two sub-agents: a sub-agent that interacts with the environment in an attempt to carry out a command, i.e., a \textit{worker}; and a sub-agent that issues commands to the worker that maximize the expected value of the return, i.e., a \textit{manager}. The problem of learning an optimal policy then becomes the problem of having both sub-agents learn to carry out their respective roles optimally. The key benefit of this formulation is that the worker solves a supervised learning problem which allows us to bring a portion of the RL problem's complexity into the domain of supervised learning, which---as we previously remarked---is the principal domain of ANNs.

Traditionally, commands in UDRL are given as desire-horizon pairs, i.e., $(d, h)$ with $d \in \mathbb{R}$ and $h \in \mathbb{N}$. Semantically, these commands direct the agent to achieve a return equal to $d$ in the next $h$ steps. Training in UDRL is done with the hindsight method wherein the agent is trained to predict what action it took given the current state and command $(g, h)$ where $g$ is the actual return observed. For UDRL to solve online RL reward-maximization problems, the commands being issued by the manager should demand higher and higher returns as time goes on.

Morethan units---as first described in \citet{schmidhuber2019reinforcement}---act as an additional element of the command. They serve to denote the relation between the true desired return and $d$. In their simplest form, morethan units capture a boolean relationship, i.e., ``get a return larger than this value". However, the concept of morethan units is flexible enough to instead encode a ternary, additive, multiplicate, or more complicated relationship.

\section{Experimental Setup}
\label{sec:experimental_setup}

We experiment with two reinforcement learning settings here: a simple six-armed bandit and the well-known CartPole domain \citep{barto1983neuronlike}. Our toy bandit domain is intended to give us a deep look at how upside-down reinforcement learning operates with the morethan unit in a basic deterministic tabular setting. With our CartPole domain, we hope to evaluate the learnability of commands with morethan units when working with non-linear function approximation in more mainstream reinforcement learning domains.

In our six-armed bandit domain, pulling the $i$-th arm always results in a reward of exactly $i$. If the agent wanted to maximize the return, it should thus always pull the $6$-th arm. Since we are working with UDRL, though, in both domains, we issue commands to the agent in the form $(d, h, m)$ where $d$ is the desired return, $h$ is a horizon, and $m$ is the \textit{morethan unit}: a ternary digit which denotes the desired relation between $d$ and $h$. If $m$ is set to $-1$, then the agent is commanded to obtain a return of less than $d$ in the next $h$ steps; if $m$ is set to $0$, then the agent should obtain a return of exactly $d$ in the next $h$ steps; and, finally, if $m$ is set to $1$, then the agent should obtain a return greater than $d$ in the next $h$ steps.

Our implementation of UDRL follows the one used by \citet{srivastava2019training} with some domain-specific adaptations. For the bandit setting, we exploit the single-step episodes by learning a two-layered policy network that uses a one-hot encoding of the command as input. Here our network uses ReLU activation and orthogonal initialization and is trained using SGD under a step size of $0.01$.

When acting in the bandit setting, we always issue a command with $m$ set to $1$. We record the results of these actions in a $100$-episode experienced replay buffer. To train the policy network with this buffer, we sample a batch of $16$ episodes and generate a permutation to pair each episode in the sample to another random episode in the sample. Recall that each of these samples will be in the form $b_i = ((d_i, h_i, m_i, g_i), a_i)$, where $d_i$ is a desired return, $h_i$ is a horizon (here always $1$), $m_i$ is the morethan unit (again here always $1$), $g_i$ is an observed return, and $a_i$ is an action. For two samples $b_i$ and $b_j$, we train the network to predict $a_i$ given an input $(g_i, h_i, m)$ where $m$ set to $1$ if $g_j > g_i$, $0$ if $g_j = g_i$, and $-1$ if $g_j < g_i$. This novel sampling strategy is critical here as it provides us with a non-parametric way of providing in-distribution samples to the network. For the bandit setting, we train once using a randomly permuted batch and once with the batch permuted such that each $b_i$ is matched with itself. We add $16$ exploratory episodes to the buffer and repeat this training process $16$ times for each iteration of the UDRL algorithm. We report the results of $10$ runs of this with $25\,000$ steps each in Section~\ref{sec:results}.

Our network in the CartPole setting has a single hidden layer of $32$ gated fast weights with Tanh activation and orthogonal initialization. We use Adam~\citep{kingma2014adam} to train this network under a step-size of $0.0008$ and with the other hyperparameters set as recommended in \citet{kingma2014adam}.

For the CartPole setting, we use a similar training regimen as in the bandit setting but use an unbounded buffer to which we add $5$ episodes in each iteration. Here, in each iteration, we train with $800$ batches of size $256$ for which we generate seven rather than two independent permutations (one of which matches samples to themselves as in the bandit setting). Note that the ability to generate exponentially more samples here is a key advantage of training with morethan units. We report the results of $30$ runs of the above with $500\,000$ steps each in Section~\ref{sec:results}.

\section{Results}
\label{sec:results}

\begin{figure}[ht]
    \centering
    \includegraphics[scale=0.725]{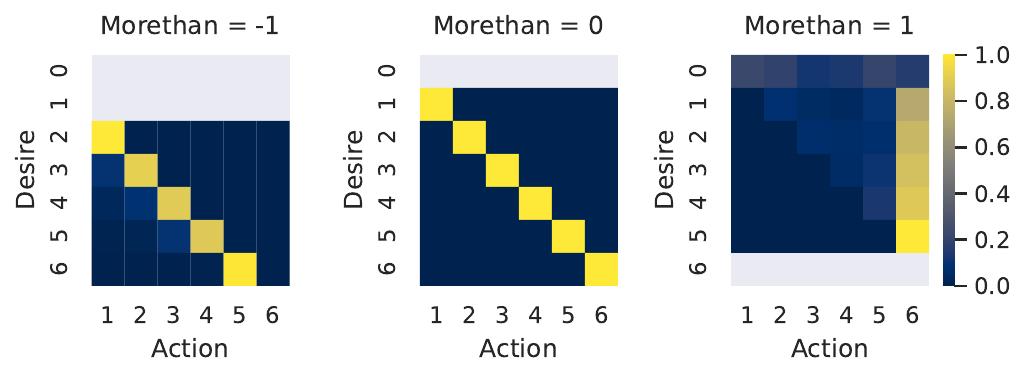}
    \caption{The mean action probabilities of the learned command-conditioned policy in the bandit setting. Note how the probability densities seem to cluster around the highest-valued valid action.}
    \label{fig:bandit_results}
\end{figure}

To evaluate learning with morethan units in the bandit setting, we look at how the action probability density of the final learned command-conditioned policy changes as a function of $d$ and $m$. The mean of these densities over the $10$ runs is shown in Figure~\ref{fig:bandit_results}. Of crucial importance here, the learned action probabilities density is concentrated in the correct regions for each setting of the morethan unit. For example, when the morethan unit is set to $0$, the action probability density is highly concentrated in the $i$-th action for each $i$ desired return. Interestingly, when the morethan unit is set to $-1$, the action probability density clusters around the $(i - 1)$-th action for each of the $i$ desired returns, but when the morethan unit is set to $1$, the action probability density is clustered around the most-rewarding action.

\begin{figure}[ht]
    \centering
    \includegraphics[scale=0.475]{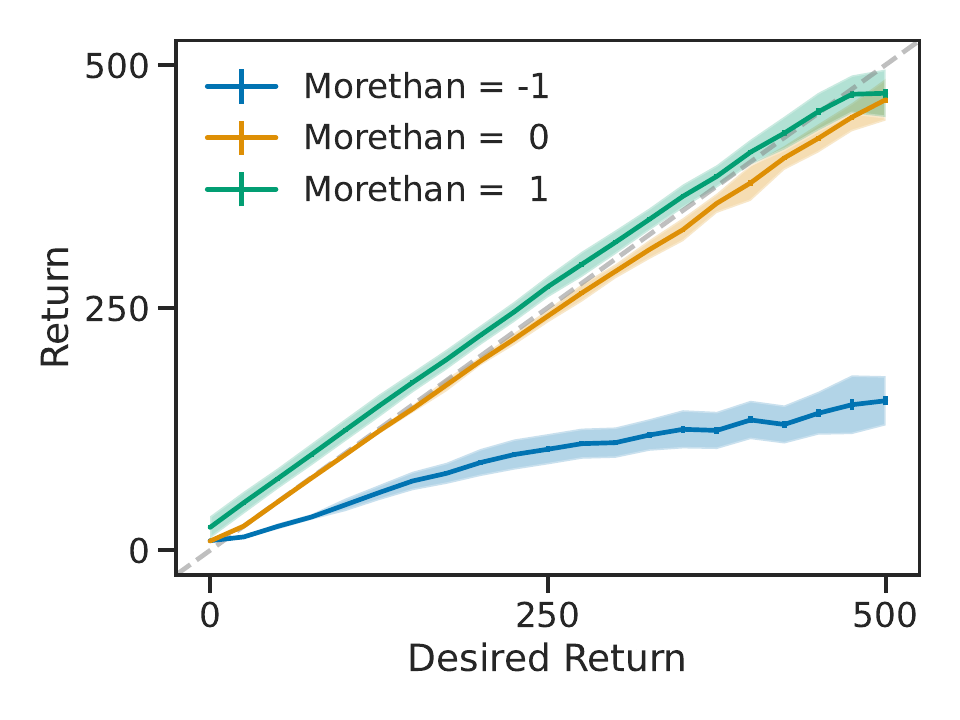}
    \caption{The desired return versus the observed return under the learned command-conditioned policy in the CartPole setting. Standard deviation is shown with shading. Note how the observed returns center around the lower-end of the valid returns.}
    \label{fig:cartpole_results}
\end{figure}

To evaluate learning with morethan units in the CartPole domain, we compare the observed return as a function of $d$ and $m$ when acting under the final learned command-conditioned policy. The results of this over the $30$ runs is given in Figure~\ref{fig:cartpole_results}. A similar but somewhat inverted relation appears here compared to the bandit results. When the morethan bit is set to $0$, the observed returns roughly correspond to the desired returns and only start to taper off slightly as the desired return reaches the higher end of the spectrum. Likewise, when the morethan bit is set to $1$, the observed return is consistently only marginally higher than the desired return for most of the values tested. However, when the morethan bit is set to $-1$, the observed returns are always drastically smaller than $d$.

\section{Discussion}
\label{sec:discussion}

This work aimed to show that morethan units were a learnable concept in the context of upside-down reinforcement learning. The results presented in Section~\ref{sec:results} clearly demonstrate that this is the case. Importantly, these results show that morethan units can be used with non-linear function approximation in mainstream online reinforcement learning domains and thus have the potential to be usable in some of the more advanced applications of reinforcement learning.

In our bandit experiments, the emphasis of the learned policy on maximal valid rewards throughout was unexpected. We hypothesize that this is due to the restrictive buffer size mixed with the training regimen whereby we perpetually demand increasingly large returns. This change in demand means that an experience replay buffer will be increasingly filled by large returns causing UDRL to increasingly oversamples large returns when training its policy network. The consequence of this is that UDRL will be prone to a specific form of forgetting. We hypothesize that this may be part of the reason that the UDRL paradigm has achieved its greatest successes primarily in the offline RL setting.

An initial inspection of the results of the CartPole experiment seems to produce something inconsistent with the above hypothesis. However, recall here that the buffer size is unbounded. This means that even though the buffer is increasingly filled with large values, the probability of a training sample having the morethan unit set to $-1$ grows as training progresses. Conversely, the probability of a training sample having the morethan unit set to $1$ shrinks as training progresses. Together with our previous hypothesis, this suggests that the distribution of training samples drastically affects the exact nature of the learned policy here, even when not otherwise affecting its validity.

\section{Conclusion and Future Work}
\label{sec:conclusion_and_future_work}

In this work, we set out to demonstrate the flexibility afforded by the use of commands in upside-down reinforcement learning. We accomplished this by experimenting with commands in the form $(d, h, m)$, where $d$ is a return, $h$ is a horizon, and $m$ is a morethan unit which denotes whether the agent is being directed to receive a return greater than, less than, or equal to $d$ in the next $h$ steps. We showed that commands of this form are learnable in two reinforcement learning environments: a six-armed bandit and the well-known CartPole environment. Through doing this, we hope to pave the way for later work to leverage advanced command structures in the hopes of successfully applying UDRL and related methods to challenging online RL problems.

Future work will look at other advanced command structures. We also plan to experiment with the morethan unit under more complicated neural network architectures---such as the Transformer architecture. This would, in turn, allow us to understand better the utility of morethan units in significantly more challenging applications.

We note that the challenges faced by the worker portion of the upside-down reinforcement learning agent are similar to the challenges faced by continual learning agents. Future work will investigate how this relationship can be leveraged to specify good command structures for UDRL and related algorithms.


\printbibliography

\end{document}